\documentclass[conference]{IEEEtran}
\IEEEoverridecommandlockouts
\usepackage{cite}
\usepackage{bm}
\usepackage{stfloats}
\usepackage{amsmath,amssymb,amsfonts}
\usepackage{algorithmic}
\usepackage{graphicx}
\usepackage{textcomp}
\usepackage{xcolor}
\usepackage{svg} 
\usepackage{color}
\usepackage{transparent}
\usepackage{graphicx}
\usepackage{import}
\usepackage{booktabs} 
\usepackage{multirow}
\usepackage{siunitx}
\usepackage{array}

\newcolumntype{L}[1]{>{\centering\let\newline\\\arraybackslash\hspace{0pt}}m{#1}}
\newcolumntype{C}[1]{>{\centering\let\newline\\\arraybackslash\hspace{0pt}}m{#1}}
\newcolumntype{R}[1]{>{\centering\let\newline\\\arraybackslash\hspace{0pt}}m{#1}}

\def\BibTeX{{\rm B\kern-.05em{\sc i\kern-.025em b}\kern-.08em
    T\kern-.1667em\lower.7ex\hbox{E}\kern-.125emX}}

\vspace{-0.8cm} 
\begin{document}

\title{Multi-source Education Knowledge Graph Construction and Fusion for College Curricula\\}
	\author{
	\thanks{*Chunhong Zhang is the corresponding author}
	\IEEEauthorblockN{
		Zeju Li,
            Linya Cheng,
		Chunhong Zhang*, 
		Xinning Zhu,
            Hui Zhao}
	\IEEEauthorblockA{Beijing University of Posts and Telecommunications}
	\IEEEauthorblockA{lizeju@bupt.edu.cn, chenglinya@bupt.edu.cn, zhangch@bupt.edu.cn,
 zhuxn@bupt.edu.cn, hzhao@bupt.edu.cn}

}
\maketitle

\begin{abstract}

The field of education has undergone a significant transformation due to the rapid advancements in Artificial Intelligence (AI). Among the various AI technologies, Knowledge Graphs (KGs) using Natural Language Processing (NLP) have emerged as powerful visualization tools for integrating multifaceted information. In the context of university education, the availability of numerous specialized courses and complicated learning resources often leads to inferior learning outcomes for students.
In this paper, we propose an automated framework for knowledge extraction, visual KG construction, and graph fusion, tailored for the major of Electronic Information. Furthermore, we perform data analysis to investigate the correlation degree and relationship between courses, rank hot knowledge concepts, and explore the intersection of courses. Our objective is to enhance the learning efficiency of students and to explore new educational paradigms enabled by AI. The proposed framework is expected to enable students to better understand and appreciate the intricacies of their field of study by providing them with a comprehensive understanding of the relationships between the various concepts and courses.

\end{abstract}


\section{Introduction}
    The integration of Artificial Intelligence (AI) into education has received significant attention due to the rapid advancements in technology and the growing demand for innovative teaching methods \cite{9839386,9853724,lihaoze2022}.
    Experts propose to utilize intelligent technology to accelerate the reform of talent training mode and teaching methods, and build a new education system including intelligent learning and interactive learning.
     For university education, each course contains a wealth of complicated learning materials, such as syllabi, textbooks, and slides, making it challenging for students to ensure quality learning outcomes and efficiency. Therefore, it is essential to explore novel and practical methods that integrate AI into university education.
    \par
    
    Knowledge graphs\cite{ji2021survey},\cite{wang2014knowledge} link distributed knowledge concepts which is scattered in various positions of the textbook together to form a large knowledge base, which describes complex relationships between entities in a structured form in the objective world. Furthermore, the application scenarios of knowledge graph are extensive, such as natural language understanding, question and answer systems based on knowledge graph, recommendation systems, and providing visual knowledge representation.
    \par

    Recently, an increasing number of researchers introduce knowledge graphs into education by the embedding knowledge for representational learning\cite{shen2021ckgg,wang2022construction}. However, there are limitations in the current research and accomplishments. One is that the human and material resources required to construct curriculum knowledge graphs are too large and the quality of knowledge extraction may be subpar. Moreover, most of the knowledge graphs constructed for the education field are small-scale graphs, or a graph of a particular chapter or a particular course. In hence, we introduce Natural Language Processing (NLP)\cite{vaswani2017attention} and knowledge graph\cite{wang2014knowledge} into university education in the field of Electronic Information. Students in the major of Electronic Information mainly learn basic circuit knowledge and master the methods of processing information with computers. We integrate the knowledge concepts of curricula and construct a visual knowledge graph of Electronic Information automatically. Students can search for knowledge concepts, view the attributes of knowledge nodes and so on. They can master the pivotal knowledge of the curriculum more efficiently and quickly. At the same time, students can also grasp the related knowledge concepts to enhance the understanding and cognition of other associated courses.

We summarize our contributions as follows:
\begin{itemize}
\item We construct multi-source and large-scale knowledge graphs for hundreds of courses in Electronic Information and import them into graph database for visualization automatically, covering more than 60k entities and 80k triples.
\item
We propose an automated knowledge extraction, data cleaning and graph fusion\cite{nguyen2020knowledge,kang2020multi} framework for heterogeneous learning resources, which also is applicable to other disciplines.
\item 
We develop auxiliary information statistics in our constructed graphs such as curriculum relevance  calculation, ranking of hot knowledge concepts, relevant keyword index of knowledge nodes and their attributes.
\end{itemize}

\section{Method}
We perform automated construction of knowledge graphs and graphs fusion can be divided as follows:
We convert textbooks and course slides from original format to editable format respectively. Next we divide the documents of textbooks and course slides into ``\textit{Knowledge Units, Knowledge Chapters, Knowledge Blocks, Knowledge Points}" automatically by using regularized expressions and named entity recognition (NER)\cite{chineseNER} methods. After that, we import chapters and knowledge points into ``Neo4j" graph database automatically to construct a large-scale knowledge graph.
In order to improve the quality of our graphs, we utilize language error correction model
\cite{2022The} enabled by deep learning for data cleaning.
Finally, we develop momentous information statistics in our constructed graphs.
        
\subsection{Ontology Construction of Subject Knowledge Graph}

Ontology construction\cite{guarino2009ontology} guides the construction of unified knowledge structure with strong hierarchy and low redundancy. In order to construct an ontology structure that can reflect the knowledge level of the college curricula and the core concepts of the knowledge content, we define classes and class inheritance. A class contains many knowledge entities and different classes imply different levels or granularity. We analyze the knowledge hierarchy division of teaching resources and design the class division of knowledge entities to obtain ontology classes and attributes. Then we define the relation type between entities to obtain the ontology relation to represent the knowledge associations structurally.

\subsubsection{Subject knowledge graph entity type definition}

An entity type is an abstraction of a set of entities that share the same characteristics or attributes and is used to distinguish the course content and knowledge granularity. Since the knowledge system was processed according to different levels of granularity based on discipline rules, we take the hierarchical catalog of teaching resources as the main basis for the construction of our knowledge graph ontology. To be specific, We start with the coarsest grain of knowledge, and then refine gradually according to the catalog level of teaching resources, to determine the maximum granularity (\textit{KnowledgeUnit}) and minimum granularity (\textit{KnowledgePoint}) of our knowledge system.

$KG_{textbook}$ indicates the knowledge graph from the textbook, $KG_{slide}$ indicates the knowledge graph from the class handout, and $KG_{syllabus}$ indicates the knowledge graph from the course syllabus. 
 On $KG_{textbook}$ and $KG_{slide}$, we further refine the knowledge system of each discipline into 5-level structure:
 \textit{Course}-\textit{KnowledgeUnit}-\textit{KnowledgeChapter}-\textit{KnowledgeBlock}-\textit{KnowledgePoint}.  \textit{KnowledgePoint} is concept-level knowledge with fine granularity, while others are course-level knowledge. On $KG_{syllabus}$, knowledge granularity was classified as ``\textit{Course}-\textit{TeachingContent}-\textit{KnowledgePoint}".

\subsubsection{Subject knowledge graph entity attribute definition}

The definition of entity attribute is a supplement to the semantic information of entity, and its coverage and accuracy affect the accuracy of the representation information of the knowledge graph. Different types of ontology attributes can enhance class information and facilitate differentiation between classes. Therefore, we design corresponding attribute sets for each type of knowledge entity. These attributes present comprehensive and specific information about the entity.

For example, We can find in Table \ref{table1} that four basic attributes (name, ranker, level, url) are set for \textit{KnowledgeUnit}, \textit{KnowledgeChapter}, \textit{KnowledgeBlock} and \textit{TeachingContent}. For the entity \textit{KnowledgePoint}, we set seven basic attributes (name, ranker, start, word\_frequency, url, level, description). The attribute ``description" can be divided into three categories: description\_wikiE, description\_wikiC and description\_baidu.

\begin{table}[ht]

  \begin{center}
    \footnotesize
    \caption{Entity types and corresponding attributes}
    \begin{tabular}{|L{2.7cm}|C{3cm}|R{1.4cm}|} 
      \hline
      
      \textbf{Entity Type} & \textbf{Entity Attributes}&\textbf{Attributes Number}\\
      \hline
      
      \textbf{Course} & school\_term, name, background,
      url, coursePrerequisites,
      educationalAlignments &6\\
      \hline
       \textbf{KnowledgeUnit} & name, ranker, level, url &4\\
      \hline
       \textbf{KnowledgeChapter} & name, ranker, level, url &4\\
      \hline
       \textbf{KnowledgeBlock} & name, ranker, level, url &4\\
      \hline
       \textbf{TeachingContent} & name, ranker, level, url &4\\
      \hline
       \textbf{KnowledgePoint} & name, ranker, start, word\_frequency, url, level, description\_wikiE, description\_wikiC, description\_baidu& 9\\
      \hline

    \end{tabular}
    \label{table1}
  \end{center}
\end{table}

\subsubsection{The entity relation definition of subject knowledge graph}

Entity relationship embodies the inner relation between different entity types. For knowledge systems, it is crucial to explore the correlation between knowledge entities, which helps students to form a deeper cumulative understanding of the curriculum knowledge system and provides a clear knowledge structure for teachers to carry out curriculum reform. Therefore, in our curriculum knowledge system, knowledge entities are interdependent. We establish knowledge correlation through the appearing chapters of teaching resource and hierarchical structure of each knowledge entity.

$KG_{textbook}$,$KG_{slide}$ and $KG_{syllabus}$ all have a relationship named ``\textit{haspartOf}" to represent knowledge granularity and hierarchy. The entity category of node links from top to bottom on $KG_{textbook}$ and $KG_{slide}$ is $\textbf{[}\ Course\xrightarrow{\textit{haspartOf}} KnowledgeUnit\xrightarrow{\textit{haspartOf}} KnowledgeChapter\xrightarrow{\textit{haspartOf}} KnowledgeBlock\xrightarrow{\textit{haspartOf}} KnowledgePoint \ \   \textbf{]}$.

Analogously, on $KG_{syllabus}$, link $\textbf{[}\ Course\xrightarrow{\textit{haspartOf}} TeachingContent\xrightarrow{\textit{haspartOf}} KnowledgePoint  \ \  \textbf{]}$ to indicate the top-down knowledge system.


\subsection{Data Cleaning}

The quality of knowledge concepts determines the quality of our constructed graphs, the process of graph fusion and the learning effect of students. In hence, data cleaning is an imperative step to guarantee the quality of our knowledge graphs. It occurs text conversion errors, format encoding and other problems when we construct knowledge graphs. And there are entity anomalies in entities and their attributes, such as text mixed with this incomplete formula and symbols, mixed Chinese and English, misspellings, syntax error of entities. Moreover, because of the algorithm limitations of the NER model and regularized expressions for knowledge concepts extraction, it may lead to syntax error in the extracted entities. 
Based on the above, we have to perform data cleaning on the original graphs we constructed. Data cleaning is a momentous step to ensure the accuracy of entities, their attributes and the effect of graph fusion because uncleaned data may lead to wrong decisions and unreliable data mining.
We use the language error correction model to check and identify the location in the entity where the incorrect word occurs. For the suspected error words, the corrected results are ranked according to the language model to select the best result, and then correct the error automatically.

\subsection{Knowledge Graph Fusion}



\begin{figure}[htbp]
    \centering
    \includegraphics[width=4cm]{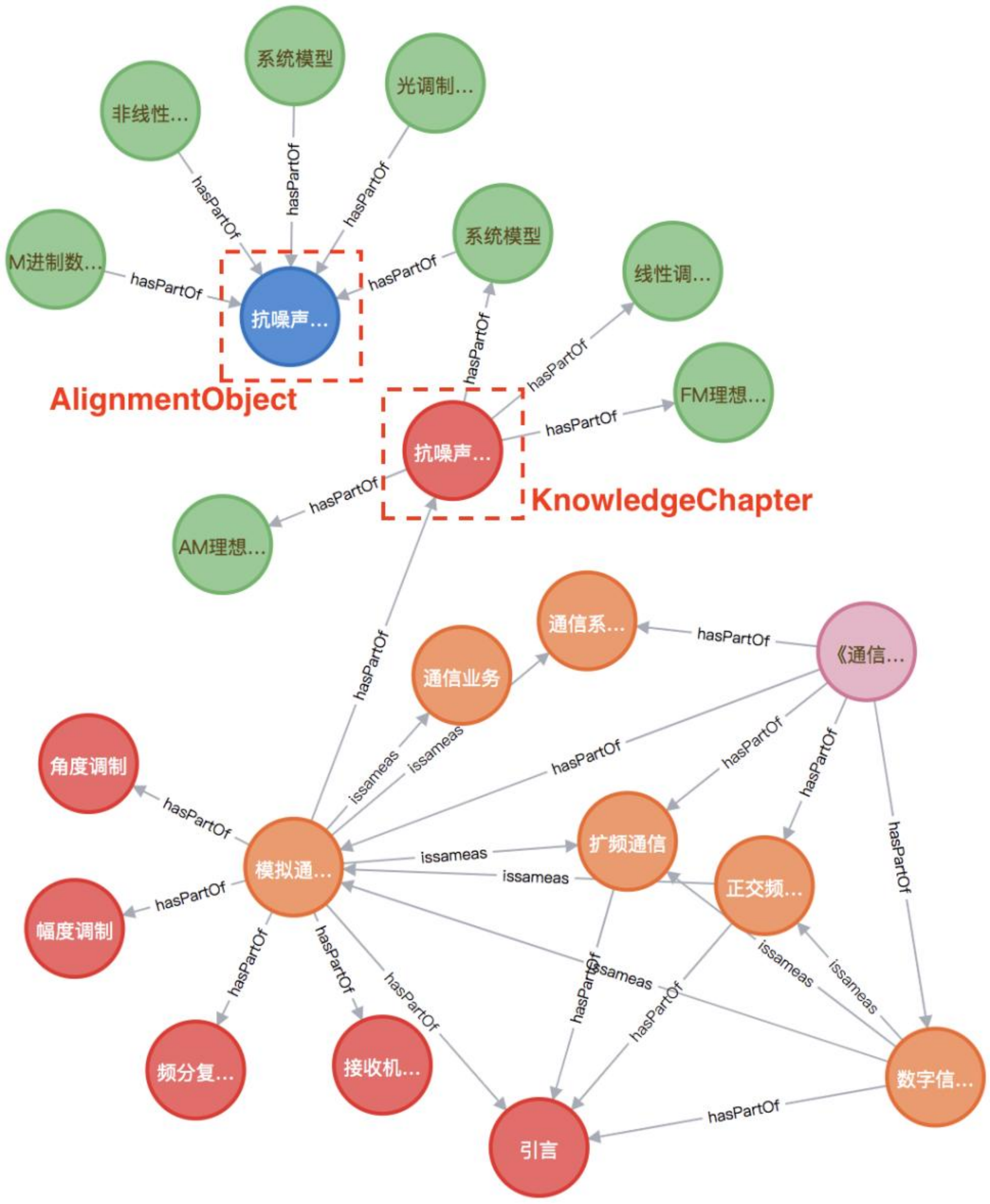}
    \caption{This is an example of KG fusion. The highlighted node is the node with the same name. In the diagram, the pink node represents the course, the orange node represents KnowledgeChapter, the red node represents Knowledgechapter, the green node represents KnowledgeBlock, and the blue node represents the AlignmentObject level, which is the lowest knowledge point.}
    \label{fig:1}
\end{figure} 

We carry out knowledge fusion to alleviate the redundant knowledge nodes and improve the data quality for our knowledge graph. Meanwhile, in order to protect the content diversity of knowledge concepts, entities referring to the same knowledge concept will be fused. Their attributes and connection relationships are reserved to maintain the information content of knowledge graph. Our main fusion method can be divided into four parts: \textit{entity name matching}, \textit{locating coarse-grained entities}, \textit{attribute union}, and \textit{preserving original edges} as follows:

\textbf{a. Entity Name Matching}

First, we match entity names to determine whether the two or more entities refer to the same knowledge concept. We used string matching, literal similarity and semantic similarity to name matching and subsequent integration.As shown in Fig.\ref{fig:1}. The first step of knowledge graph fusion is to integrate these fragmentary nodes referring to the same knowledge into a unique node.

\textbf{b. Locating Coarse-grained Entities}

As described in the previous section, knowledge concepts (entities) often exist at different levels in the knowledge system of textbooks. The higher the level, the larger the coverage and the coarser the granularity. We define the entities to be fused as $E=\{e_1,e_2,....,e_n\}$ and their levels are $L=\{l_1,l_2,...l_n\}$. For example, $e_f$ is used to represent the fused Knowledge node (entity), $l_f$ represents the unique level of $e_f$. In order to more completely retain the hierarchical structure in the knowledge graph, $e_f$ is located at the coarsest node in $E$ and $l_f=highest(L)$.

\textbf{c. Taking Union of Attributes}

In order to maintain the amount of information of the knowledge graph after the knowledge fusion work, in addition to retaining the hierarchical relationship, we also need to retain the union operation for the attribute information (except ``url") of each knowledge concept (entity) before fusion.

We define attribute set $A=\{A_1,A_2,...,A_n\}$ to represent all attribute information corresponding to entity set $E$ to be fused. $A_1=\{...,a_1,b_1,...\}, A_2=\{...,a_2,b_2,...\}, A_i=\{...,a_i,b_i,...\}$, where $a_i$ represents the $a$ property of entity $e_i\in E$, and $b_i$ represents the $b$ property of entity $e_i\in E$. $A_f$ represents the attribute set of the node $e_f$ after fusion, which can be expressed as:
$$A_f= A_1\cup A_2...\cup A_n =\{...,\sum_{i=1}^{n}a_i,\sum_{i=1}^{n}b_i,.....\}$$
For example, the two ``anti-noise performance" knowledge concepts have the attributes shown in Fig.\ref{fig:1} respectively, then the unique knowledge concepts obtained after the fusion work had the union of the attributes of the two nodes, that is, the attributes of name, description, word\_frequency and ranker, so the attribute information is most reserved.

\textbf{d. Preserving Original Edges}

In different textbooks, the topological structure and connection relationship between knowledge nodes (entities) are different. In addition to the retention of entity information, relational information should also be processed accordingly to achieve information retention. Since the only node refer to one concept in the textbook will be obtained after fusion work, all the connection relationships existing before the fusion work should exist and be connected to the unique knowledge concept after the fusion work.

Our graph fusion types are divided into different source fusion and different course fusion. We utilize the methods above for graph fusion of the same course. Moreover, we utilize entity attribute matching, preserving original edges and creating equivalent edges for graph fusion between courses. 

\section{Experiment and Analysis}

\subsection{Analysis of the correlation degree between two courses}

In this part, we judge the degree of correlation between any two courses by calculating the proportion of the number of equivalence relation edges  between two courses in the total knowledge concepts. First of all, we integrate within the curriculum and the same course. Under the condition that the original knowledge graph structure system remains unchanged. According to the result of fusion, we propose that there was a similar relationship between courses, so we proposed the index of correlation degree between courses, which was defined by the proportion of the number of knowledge concepts between courses and the total nodes of the course.


We check the coverage of knowledge concepts between courses by counting the number of equivalent relation edges established between two courses. $KG_{textbook}=\{... V_i, V_j,.... \}, KG_{slide} = \{... U_i, U_j,.... \}$, $V_i$ represents the course in the knowledge graph of the textbook, and the total number of nodes of this course is $Nv_i$; $U_i$ represents the course in the slide knowledge graph, and the total number of course nodes is $Nu_i$. The number of mapping equivalent edges established between course $V_i$ and course $V_j$ is expressed as $sim\_v_{i,j}$, and the number of mapping equivalent edges established between course $U_i$ and course $U_j$ is expressed as $sim\_u_{i,j}$. The knowledge graph of the two data sources is fused respectively, and an overall course correlation degree is defined as $S$:
$$
S(i,j)=\frac{Su(i,j)+Sv(i,j)}{2}
$$
The correlation degree calculated by the slide database is the correlation degree $Su$, and the correlation degree calculated by the textbook database is the correlation degree $Sv$. We comprehensively analyzed the data results by combining the data from two data sources and took them as the average total correlation degree of the correlation degree of the two courses. $Su$ and $Sv$ are calculated as follows:

$$Su(i,j)=\frac{sim\_u_{i,j} }{ Nu_i}, Sv(i,j)=\frac{sim\_v_{i,j}} { Nv_i}$$

\begin{table*}[ht]
  \begin{center}
    \scriptsize
    \caption{Calculating result of the correlation degrees with ``\textit{Communication Principles}''}
    \begin{tabular}{|L{5cm}|L{5cm}|L{1.5cm}|L{1.5cm}|L{1.5cm}|} 
      \hline
      
      \textbf{$Course_i$} & \textbf{$Course_j$} &\textbf{$S_u(i,j)$} &\textbf{$S_v(i,j)$} &\textbf{$S(i,j)$}\\
      \hline
      
      \textit{Communication Principles} & \textit{Signals and Systems} &0.11389 &0.14008&0.12698\\
      
      \hline
      \textit{Communication Principles} & \textit{Communication Electronic Circuit} &0.11676 &0.05589&0.08632\\

      \hline
      \textit{Communication Principles} & \textit{Fundamentals of Information Theory } &0.09912 &0.05589&0.07751\\
      \hline
      \textit{Communication Principles} & \textit{Fundamentals of Probability Theory and Mathematics} &0.06768 &0.05078&0.05923\\
      \hline
      \textit{Communication Principles} & \textit{Linear Algebra and Geometry} &0.08368 &0.01761&0.05065\\
      \hline
      \textit{Communication Principles} & \textit{Digital Circuits and Logic Design} &0.2265 &0.05723&0.03994\\
      \hline
      \textit{Communication Principles} & \textit{Electronic and Circuit Foundation} &0.02702 &0.04336&0.03519\\
      \hline
      \textit{Communication Principles} & \textit{Digital Signal Processing} &0.01923 &0.04813&0.03368\\
       \hline
      \textit{Communication Principles} & \textit{C/C++ Programming} &0.04177 &0.02548&0.03362\\
       \hline
      \textit{Communication Principles} & \textit{Electromagnetic Field and Electromagnetic Wave} &0.03956 &0.00998&0.02477\\
       \hline
      \textit{Communication Principles} & \textit{University Physics} &0.03242 &0.00540&0.01891\\
      \hline

    \end{tabular}
    \label{table2}
  \end{center}
\end{table*}

\begin{table}[ht]
  \begin{center}
    \scriptsize
    \caption{The most relevant courses with each course and corresponding correlation degrees}
    \begin{tabular}{|L{3cm}|L{3cm}|L{1cm}|} 
      \hline
      
      \textbf{Course} & \textbf{Most Relevant Course} &\textbf{Correlation Degree} \\
      \hline
      
      \textit{Communication Principles} & \textit{Signals and Systems} &0.127 \\
      \hline
      \textit{Signals and Systems} & \textit{Communication Electronic Circuit} &0.119 \\
      \hline

      \textit{Digital Circuits and Logic Design} & \textit{Electronic and Circuit Foundation} &0.201 \\
      \hline
      \textit{Fundamentals of Probability Theory and Mathematics} & \textit{Linear Algebra and Geometry} &0.084 \\
      \hline
      \textit{Electronic and Circuit Foundation} & \textit{Communication Electronic Circuit} &0.345 \\
      \hline
      \textit{Electromagnetic Field and Electromagnetic Wave} & \textit{University Physics} &0.029 \\
      \hline
      \textit{Linear Algebra and Geometry} & \textit{Fundamentals of Information Theory} &0.015 \\
      \hline
      \textit{Digital Signal Processing} & \textit{Signals and Systems} &0.064 \\
      \hline
      \textit{University Physics} & \textit{Electromagnetic Field and Electromagnetic Wave} &0.101 \\
      \hline
      \textit{C/C++ Programming} & \textit{Linear Algebra and Geometry} &0.046 \\
      \hline
      \textit{Communication Electronic Circuit} & \textit{Signals and Systems} &0.035 \\
      \hline
      \textit{Fundamentals of Information Theory} & \textit{Communication Principles} &0.149 \\
      \hline
    \end{tabular}
\label{table3}
  \end{center}
\end{table}

Taking the course of \textit{Communication Principles} as an example, the result of calculating the correlation degree with the course is shown in Table \ref{table2}.
 The most relevant courses of other courses in Electronic Information and their corresponding correlation degrees are shown in Table \ref{table3}. we can see the courses that are highly related to \textit{Communication Principles}. In the actual teaching application scenario, it can be shown that these courses with high correlation can play an auxiliary role in the study of \textit{Communication Principles}, which helps teachers to arrange appropriate teaching programs and enables students to have a macro overall control over the relationship between the courses they need to learn.
The method we proposed has great advantages in comparing the similarity between courses, because we not only consider the correlation between the two course names, but also consider the overlap rate of knowledge concepts at each level of the course, so as to determine the relationship between courses.

\subsection{Explore the relationship between multiple courses }

According to the method of calculating the correlation degree of two courses, we can sort each subject with other subjects in the order of correlation degree, so as to analyze the relationship between multiple courses and classify them.

We used the spectral clustering method\cite{von2007tutorial} and input the correlation degree matrix, and the results obtained are highly similar to the results of the manual classification mentioned above, which can be used as the clustering method of course classification. 
Firstly, we choose ten subjects with relatively complete data resources, and the set of courses supported by data is $C$, $C=\{c_1,c_2,...c_i,... c_n\} (n=10)$. Correlation degree $S(i, j)$ represents the degree of correlation between $c_j$ and $c_i$. So for every $c_i$ course, there's $S_w(i)=\{S(i,1),S(i,2),..... ,S(i,j),... \}$, represents $c_i$'s set of correlation degrees with other courses. When the correlation comparison of multiple courses is needed, we propose $W(i,j)$ to represent the correlation weight of $c_j$ to $c_i$. Therefore, each course $c_i$ has correlation with other courses set weight $W(i) = \{W (i, 1), W(i, 2),... ,W(i,j),... \}$. In the same course, we normalize the correlation degree to get the normalized correlation degree of each $c_j$ of other courses about the $c_i$ of this course, which can better represent the degree of correlation in the same dimension.

\begin{figure}[htbp]
\begin{minipage}[p]{0.45\linewidth}
\centering
\includegraphics[height=3cm,width=4.35cm]{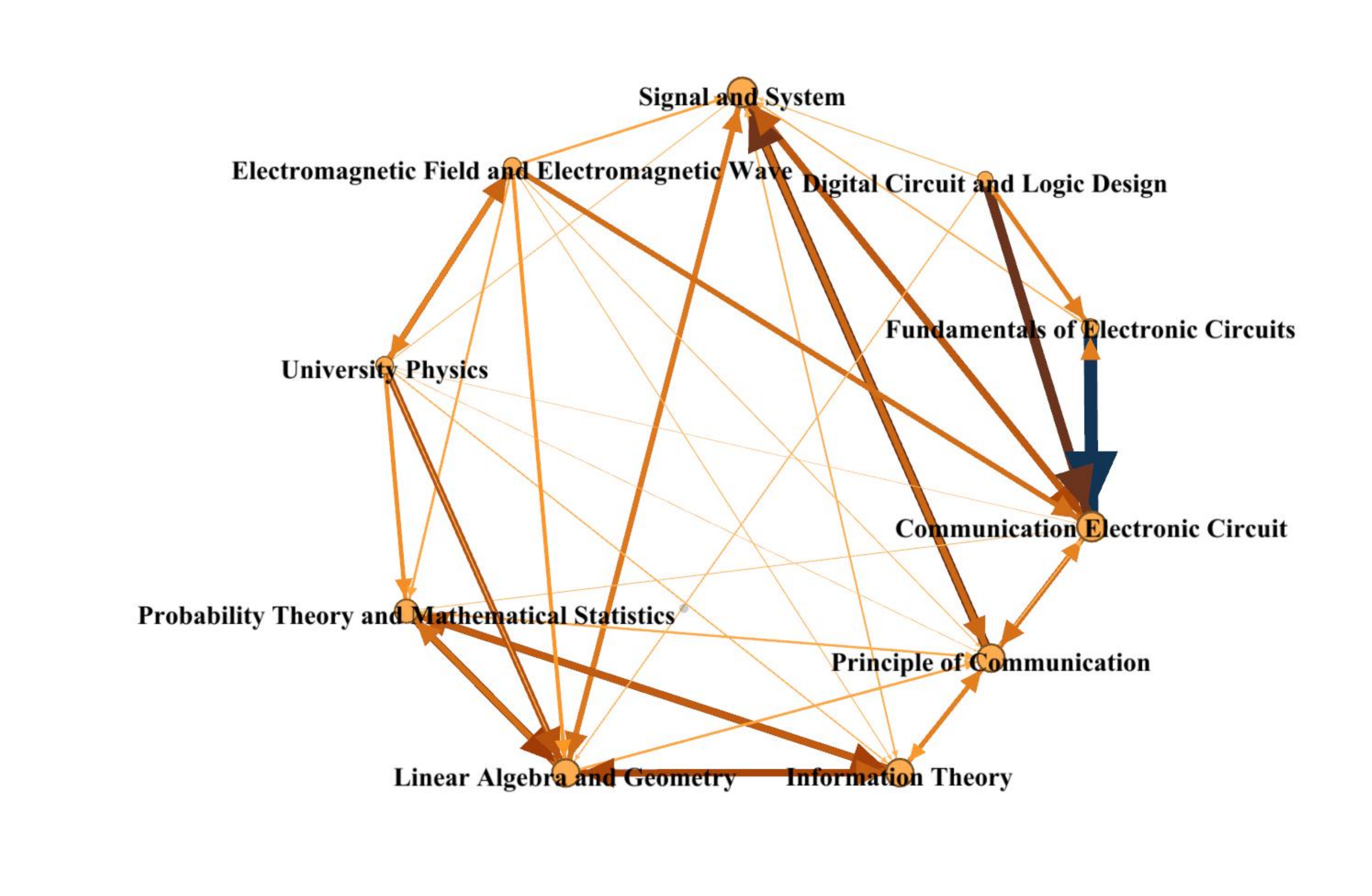}
\caption{Graph of correlation degree weight of multiple courses}
\label{fig:2}
\end{minipage}%
\begin{minipage}[p]{0.45\linewidth}
\centering
\includegraphics[height=3.4cm,width=3.1cm]{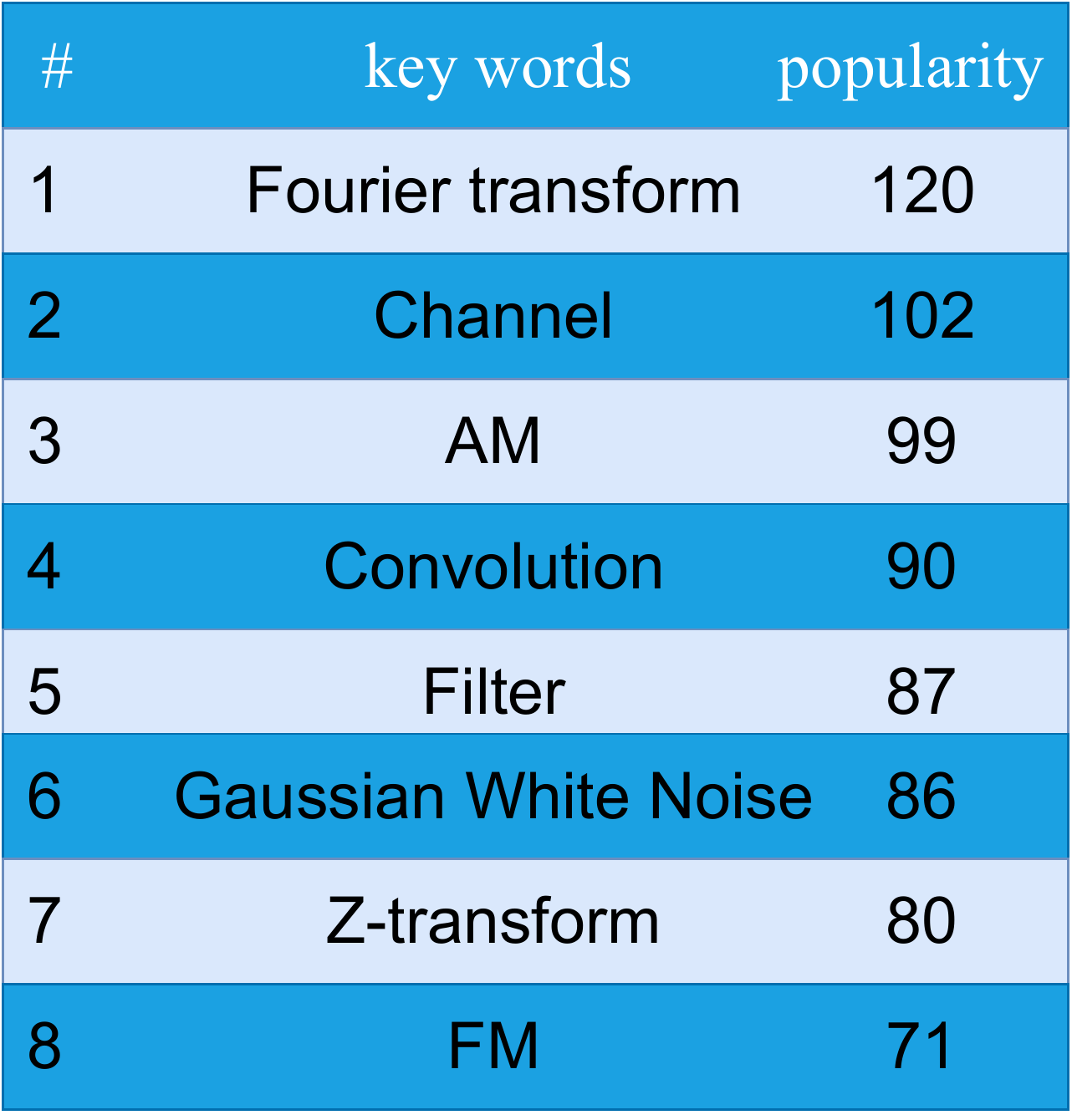}
\caption{The ranking of the knowledge concepts frequency in the interface of our KG visual system.}
\label{fig:3}
\end{minipage}

\end{figure}

In order to visualize the relationship more intuitively between multiple courses, we use the thickness of the correlation edge to reflect the weight of the correlation degree. In order to facilitate the observation of the correlation degree between courses, we remove the correlation edge with the weight less than 0.25, as shown in the Fig.\ref{fig:2}. The results of the relationship between multiple courses obtained in this part are of certain reference value for teachers to make teaching plans.

\subsection{Explore the most valuable and integral knowledge concepts}
Since our knowledge graphs contain massive knowledge concepts, 
it is difficult for students to focus on the most valuable and significant knowledge concepts among them. In hence, we separately count the number of occurrences and calculate the intersection of all knowledge concepts in the database.
Firstly, we count the frequency of occurrences of all knowledge concepts. We can obtain the ranking of the knowledge concepts frequency in Fig.\ref{fig:3}, so we can explore the most significant knowledge concepts in Electronic Information.

And then, we count the intersection of knowledge concepts of the same course among its $KG_{textbook}$, $KG_{slide}$ and $KG_{syllabus}$ after the fusion of graphs.
In this way, we can obtain the most valuable knowledge concepts in one course, because they appear in $KG_{textbook}$, $KG_{slide}$ and $KG_{syllabus}$ at the same time. We can find that 16 knowledge concepts in \textit{Communication Principles} appear in $KG_{textbook}$, $KG_{slide}$ and $KG_{syllabus}$ :
``inner product, orthogonality, correlation coefficient, energy power spectral density, Hilbert transform, analytic signal, band-pass signal, matched filter, FDM, TDM, Nyquist criterion, raised cosine roll-off, phase ambiguity, optimum Reception, constellation, spectrum".
In this way, We can find the most valuable knowledge concepts in \textit{Communication Principles}. So teachers and students will focus on these knowledge concepts when teaching and learning.


\section{Discussion And Conclusion}
We propose a novel method of knowledge graph construction and graphs fusion by means of natural language processing and machine learning tools, which is a automatic framework for heterogeneous learning resources and is applicable and suitable to other disciplines.
Through our graph fusion method, the individual knowledge graph is fused into the knowledge graph from multiple data sources. Meanwhile, through the data mining and analysis, we infer the relationship between courses in Electronic Information and create the relationship between disciplines to sort out the discipline knowledge framework.

Our method not only conforms to the cognitive habits of learners, but also improves learning efficiency by helping students to understand the knowledge concepts between various courses and  grasps integral knowledge concepts of one certain subject comprehensively. 
Moreover, it can provide powerful data support for teachers to achieve the corresponding teaching objectives and highlight the core knowledge concepts to improve teaching quality.


\section*{Acknowledgment}
This work is supported  by 2022 Beijing Higher Education ``Undergraduate Teaching Reform
and Innovation Project" and 2022 Education and Teaching Reform Project of Beijing University of Posts and Telecommunications (2022JXYJ-F01).

\bibliographystyle{IEEEtran}
\bibliography{conference_101719}
\end{document}